\documentclass{article}
\usepackage{spconf,amsmath,amssymb,graphicx,booktabs,hyperref}
\def\x{{\mathbf x}}
\def\y{{\mathbf y}}
\def\F{{\mathbf F}}
\def\C{{\mathbf C}}
\def\M{{\mathbf M}}

\def\R{{\mathbf R}}
\title{FreePath: Neural Mean Free Path for Image Dehazing}

\name{Rongrong Gao$^{1}$ \qquad Jingxuan Feng$^{2}$ \qquad Wenyu Xu$^{3}$}
  
  \address{$^{1}$ Tsinghua University $^{2}$Wuchang University of Technology $^{3}$The Hong Kong Polytechnic University}

\begin{document}
\maketitle

\begin{abstract}
Single image dehazing requires recovering scene radiance from visual responses
that are weakened and veiled by participating media. Existing methods either
invert explicit scattering variables, which can be unstable under complex
scenes, or learn direct restoration mappings, which often discard the physical
organization of haze formation. We present \emph{FreePath}, a neural mean free
path framework for image dehazing. Instead of estimating atmospheric
parameters, FreePath represents dehazing as feature space path state
restoration, inspired by the expected distance a visual response can propagate
before interacting with haze. Within a residual encoder and decoder, the Mean
Free Path Residual Block learns path and collision fields to guide haze
constrained correction, while FreePath State Conservation Integration preserves
restored path states during deep feature aggregation. Experiments on paired
CDD-11 haze images and real RTTS scenes show that FreePath consistently
improves full reference fidelity, no reference perceptual quality, and visual
faithfulness over representative dehazing baselines.
\end{abstract}


\section{Introduction}
\label{sec:intro}

Images captured in haze lose contrast, color fidelity, and distant structures,
which makes visual understanding and downstream perception unreliable
\cite{narasimhan2002vision,he2010single}. Classical dehazing
explains this degradation through atmospheric scattering, where the recorded
image mixes attenuated scene radiance with airlight scattered by suspended
particles \cite{koschmieder1925theorie,narasimhan2002vision}. This view has
inspired influential priors and neural designs, but the usual variables are
often too coarse to describe the feature ambiguity handled by learned
restoration networks \cite{he2010single,meng2013efficient,berman2016non,li2017aod}.

Learning based dehazing improves restoration through scattering related
prediction, gated or attention based aggregation, and general restoration
architectures \cite{li2017aod,chen2019gated,qin2020ffa,zamir2022restormer}. Yet
dehazing still needs a representation that is more physically meaningful than
generic enhancement and less brittle than explicit single image parameter
estimation \cite{ishimaru1999wave,bohren2008absorption,zheng2023curricular}.
From this perspective, a dehazing network should identify which feature
responses can remain as reliable scene evidence and which responses require
medium aware correction. This motivates a learnable path state that connects a
physical notion of path freedom with residual image restoration.

We address this gap with \emph{FreePath}, a neural mean free path framework.
Mean free path describes the expected travel distance before a medium
interaction \cite{ishimaru1999wave,bohren2008absorption}. FreePath translates
this concept into a feature space path state, where path free responses preserve
scene structure and collision constrained responses receive stronger residual
correction.

\begin{figure*}[t]
    \centering
    \includegraphics[width=0.98\textwidth]{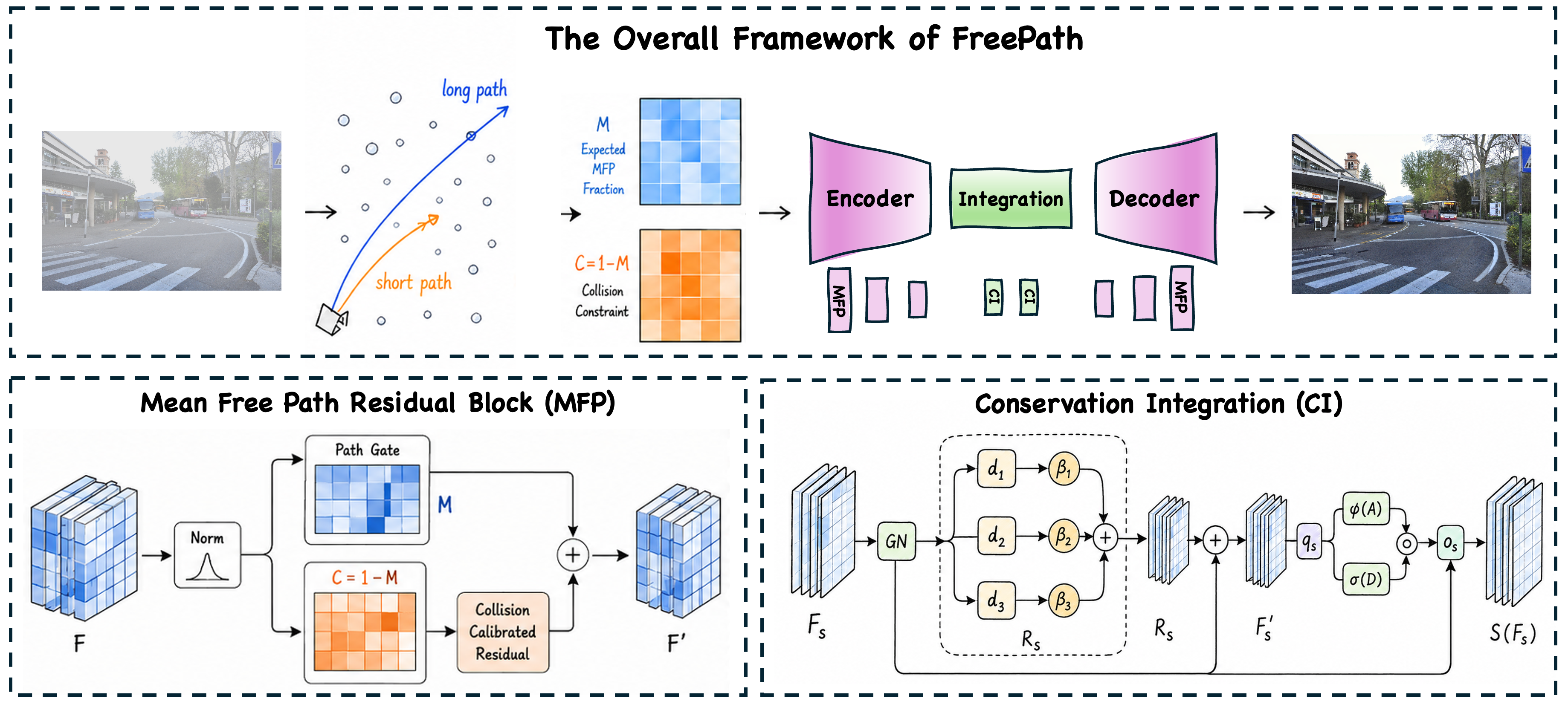}
    \vspace{-10pt}
    \caption{The overall framework of FreePath. The first row shows neural mean
    free path restoration inside an encoder and decoder pipeline. The
    lower panels show the Mean Free Path Residual Block and Conservation
    Integration, which perform path guided correction and state preserving
    aggregation.}
    \label{fig:main}
\end{figure*}

As shown in Fig.~\ref{fig:main}, FreePath keeps the residual encoder and
decoder topology but reorganizes internal computation around this path state.
The Mean Free Path Residual Block corrects hazy responses, and FreePath State
Conservation Integration consolidates deep features without assigning another
path state. Our contributions are summarized as:
\begin{itemize}
    \item We propose FreePath, a neural mean free path framework that
    reformulates image dehazing as feature space path state restoration.
    \item We instantiate the framework with a Mean Free Path Residual Block
    and a FreePath State Conservation Integration module, respectively
    correcting collision constrained responses and preserving restored path
    states during deep aggregation.
    \item Experiments on paired CDD-11 haze images and real RTTS scenes
    demonstrate strong full reference accuracy, no reference quality, and
    visually faithful dehazing performance.
\end{itemize}

\section{Related Work}
\label{sec:related}

\noindent\textbf{Physical image dehazing.}
Physical dehazing methods are rooted in atmospheric scattering and radiative
propagation \cite{ishimaru1999wave,bohren2008absorption,
narasimhan2002vision}. Following this formulation, classical algorithms design
image priors based on contrast maximization, dark channel statistics, boundary
constraints, color attenuation, haze lines, and probabilistic inference
\cite{fattal2008single,he2010single}. These
methods expose meaningful physical structure, but their hand crafted
assumptions can be fragile when haze density, illumination, and scene content
vary jointly \cite{li2018benchmarking,ancuti2019dense,guo2024onerestore}.

\noindent\textbf{Learning based dehazing and restoration.}
Learning based dehazing has benefited from synthetic and real benchmarks
\cite{li2018benchmarking,ancuti2018haze, guo2024onerestore}. Early networks estimate haze related maps or directly
regress clear images \cite{cai2016dehazenet,ren2016single,li2017aod}, while later methods introduce gated fusion, adversarial
translation, grid processing, feature attention, contrastive learning, domain
adaptation, codebook priors, and transformer architectures
\cite{ren2018gated,chen2019gated,wu2023ridcp}. The evaluated baselines span gated context aggregation
\cite{chen2019gated}, rank oriented recovery \cite{liu2022rank}, all in one
restoration \cite{li2022all}, prompt driven adaptation
\cite{zhang2026pdda}, and haze density correction \cite{liu2026ihdcp}. General
restoration models further motivate stable encoder and decoder restoration
\cite{zamir2021multi,zamir2022restormer,liang2021swinir,
wang2022uformer,potlapalli2023promptir}, while FreePath organizes this restoration with a
neural mean free path state.

\section{Method}
\label{sec:method}

\subsection{Overview}

FreePath is designed to keep the optimization reliability of a
residual encoder and decoder while giving its feature evolution a scattering
aware organization. Given a hazy image
$\x \in \mathbb{R}^{3 \times H \times W}$, the network predicts a dehazed
image $\hat{\y}$ as
\begin{equation}
    \hat{\y} = \x + \Phi_{\mathrm{out}}
    \bigl( D( S( E(\Phi_{\mathrm{in}}(\x)) ) ) \bigr),
\end{equation}
where $\Phi_{\mathrm{in}}$ and $\Phi_{\mathrm{out}}$ are $1 \times 1$
projection layers, $E$ is a four stage encoder, $S$ is the FreePath State
Conservation Integration module, and $D$ is the decoder with skip connections.
The residual output form encourages the model to focus on haze removal rather
than reconstructing the whole scene from scratch, which is consistent with
recent restoration networks
\cite{zamir2021multi,zamir2022restormer,chen2022simple}.

The key representation in FreePath is the neural mean free path state. Each
Mean Free Path Residual Block estimates a path gate $\M$ and a complementary
collision constraint $\C=1-\M$. The gate is not treated as a calibrated
physical distance. It is a learned feature field whose role mirrors the mean
free path concept: large $\M$ marks responses that can be carried forward as
scene evidence, while large $\C$ marks responses that should receive stronger
haze correction \cite{ishimaru1999wave,bohren2008absorption}. The integration
stage then aggregates deep features without creating a second path assignment,
so the restored state remains consistent between the encoder and decoder.

\subsection{Mean Free Path Residual Block}

The Mean Free Path Residual Block is the core computation unit of FreePath. For
an input feature $\F$, the block first normalizes the feature and predicts the
path state:
\begin{equation}
    \M = \sigma(g(\mathrm{GN}(\F))), \qquad \C = 1 - \M,
\end{equation}
where $g(\cdot)$ is a convolutional predictor and $\sigma$ denotes
the sigmoid function. The normalized feature is modulated by the free path
field before residual extraction:
\begin{equation}
    \tilde{\F} = (0.5 + \M) \odot \mathrm{GN}(\F).
\end{equation}
This keeps path free responses prominent and assigns larger residual correction
to collision constrained responses.

To retain the strong optimization behavior of convolutional restoration blocks
\cite{dong2020multi,zamir2021multi,chen2022simple}, FreePath uses a set
of depthwise residual filters and predicts data dependent response
weights:
\begin{equation}
    \R = \sum_{k=1}^{3} \alpha_k(\tilde{\F}) \, d_k(\tilde{\F}),
    \quad \sum_{k=1}^{3}\alpha_k=1,
\end{equation}
where $d_k$ are depthwise filter bases. The collision constraint controls the
residual energy:
\begin{equation}
    \F' = \F + r\bigl((0.75 + \C) \odot \R\bigr),
\end{equation}
where $r(\cdot)$ is a pointwise recomposition layer. A gated channel mixer then
produces the block output. The block therefore does not invert an explicit haze
equation. It learns residual updates from paired supervision, but the updates
are routed by the neural free path and collision fields.

\subsection{FreePath State Conservation Integration}

After repeated path sensitive correction, the deepest representation should be
aggregated without changing its recovered path state. FreePath therefore uses
State Conservation Integration. Its principle is simple: once the restoration
stream has inferred neural mean free path states, the integration stage should
consolidate deep features instead of predicting another independent path gate.

Given a deep feature $\F_s$, the integration module applies a path preserving
residual transform:
\begin{equation}
    \R_s = \sum_{k=1}^{3} \beta_k(\bar{\F}_s)\, d_k(\bar{\F}_s),
    \quad \bar{\F}_s=\mathrm{GN}(\F_s), \quad
    \sum_{k=1}^{3}\beta_k=1 ,
\end{equation}
where $d_k$ are depthwise residual filter bases and $\beta_k$ are adaptive
response weights. The integrated feature is then obtained by
\begin{equation}
    \F_s' = \F_s + r_s(\R_s),
\end{equation}
followed by a gated channel integration:
\begin{equation}
    S(\F_s)=\F_s' + o_s\bigl(\phi(\mathbf{A})\odot\sigma(\mathbf{D})\bigr),
    \quad [\mathbf{A},\mathbf{D}]=q_s(\mathrm{GN}(\F_s')) .
\end{equation}
Different from the Mean Free Path Residual Block, this module does not predict
a new free path gate or collision field. It preserves the path states restored
by surrounding FreePath blocks and performs stable deep feature aggregation.
Thus, FreePath first corrects features according to inferred path freedom and
then conserves those states for decoding.

\subsection{Training Objective}

We train FreePath with image level restoration supervision. The hazy input is
normalized to $[-1,1]$, and the prediction is mapped back to $[0,1]$ before
being compared with the clear target $\y$:
\begin{equation}
    \mathcal{L} =
    0.7\,\mathcal{L}_{\mathrm{SL1}}(\hat{\y},\y)
    + 0.3\,\mathcal{L}_{\mathrm{MSSSIM}}(\hat{\y},\y)
    + 0.01\,\mathcal{L}_{\mathrm{perc}}(\hat{\y},\y),
\end{equation}
where $\mathcal{L}_{\mathrm{SL1}}$ is the smooth $\ell_1$ loss,
$\mathcal{L}_{\mathrm{MSSSIM}}=1-\mathrm{MSSSIM}$, and
$\mathcal{L}_{\mathrm{perc}}$ is a VGG 16 feature perceptual loss computed with
the first 16 feature layers fixed \cite{wang2004image,zhang2018unreasonable}.
The mean free path and collision fields are not directly annotated; they emerge
from paired image restoration.

\begin{table*}[t]
    \centering
    \caption{Main comparison on CDD-11 and RTTS. Results are reported as mean
    $\pm$ standard deviation. Higher is better for PSNR, SSIM, FSIM, VSI, and
    CLIPIQA, while lower is better for LPIPS and GMSD.}
    \vspace{5pt}
    \label{tab:main}
    \setlength{\tabcolsep}{2pt}
    \begin{tabular}{@{}lccccccc@{}}
        \toprule
        & \multicolumn{6}{c}{CDD-11} & \multicolumn{1}{c}{RTTS} \\
        \cmidrule(lr){2-7}\cmidrule(l){8-8}
        Method & PSNR $\uparrow$ & SSIM $\uparrow$ & FSIM $\uparrow$ & VSI $\uparrow$ & LPIPS $\downarrow$ & GMSD $\downarrow$ & CLIPIQA $\uparrow$ \\
        \midrule
        GCANet~\cite{chen2019gated} & 17.991$\pm$4.500 & 0.893$\pm$0.063 & 0.940$\pm$0.034 & 0.979$\pm$0.013 & 0.154$\pm$0.081 & 0.048$\pm$0.019 & 0.377$\pm$0.210 \\
        ROP~\cite{liu2022rank} & 17.833$\pm$3.113 & 0.930$\pm$0.041 & 0.927$\pm$0.030 & 0.969$\pm$0.015 & 0.139$\pm$0.069 & 0.044$\pm$0.017 & 0.353$\pm$0.190 \\
        AirNet~\cite{li2022all} & 17.012$\pm$2.754 & 0.607$\pm$0.117 & 0.897$\pm$0.022 & 0.972$\pm$0.009 & 0.137$\pm$0.049 & 0.137$\pm$0.026 & 0.379$\pm$0.135 \\
        PDDA~\cite{zhang2026pdda} & 18.761$\pm$2.629 & 0.886$\pm$0.046 & 0.951$\pm$0.032 & 0.980$\pm$0.014 & 0.176$\pm$0.067 & 0.045$\pm$0.025 & 0.394$\pm$0.150 \\
        IHDCP~\cite{liu2026ihdcp} & 19.550$\pm$3.725 & 0.941$\pm$0.039 & 0.960$\pm$0.024 & 0.982$\pm$0.012 & 0.078$\pm$0.053 & 0.027$\pm$0.011 & 0.423$\pm$0.207 \\
        FreePath & \textbf{27.608$\pm$4.549} & \textbf{0.986$\pm$0.012} & \textbf{0.990$\pm$0.010} & \textbf{0.997$\pm$0.003} & \textbf{0.016$\pm$0.014} & \textbf{0.010$\pm$0.008} & \textbf{0.433$\pm$0.199} \\
        \bottomrule
        
    \end{tabular}
\end{table*}

\begin{figure*}[t]
    \centering
    \includegraphics[width=0.98\textwidth]{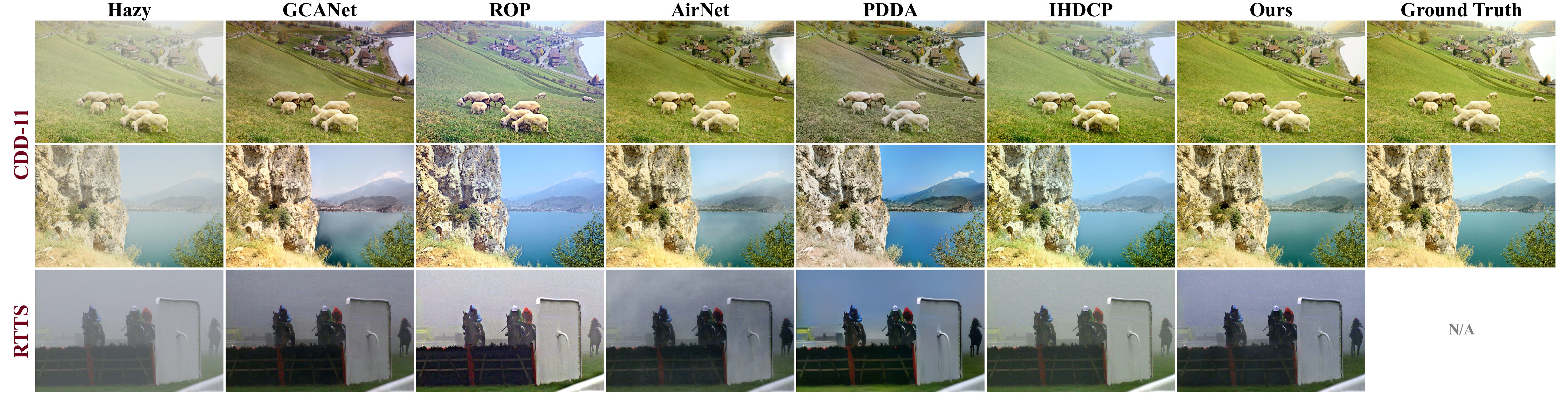}
    \vspace{-5pt}
    \caption{Visual comparison on CDD-11 and RTTS. Columns follow
    Table~\ref{tab:main}. Ground truth is available only for CDD-11.}
    \label{fig:visual}
    \vspace{-10pt}
\end{figure*}

\section{Experiments}
\label{sec:exp}

\subsection{Dataset and Protocol}
Experiments use two complementary datasets. For paired training and full
reference evaluation, we use the haze subset of CDD-11
\cite{guo2024onerestore}, which provides aligned hazy and clear images. We sort
the training pairs by numerical identifiers and train FreePath on the first
500 hazy and clear pairs. Testing uses 200 paired CDD-11 haze images. We report
PSNR, SSIM~\cite{wang2004image}, FSIM~\cite{zhang2011fsim},
VSI~\cite{zhang2014vsi}, LPIPS~\cite{zhang2018unreasonable}, and
GMSD~\cite{xue2013gradient}.

For real scene evaluation, we select 50 RTTS images from RESIDE
\cite{li2018benchmarking}. Since RTTS has no aligned clear targets, we report
CLIPIQA and visual comparisons to assess no reference dehazing quality.

\subsection{Implementation Details}
All models, including FreePath and the ablation variants, are trained under the
same setting on a single NVIDIA RTX 3080 GPU. We use Adam optimization with an
initial learning rate of $10^{-4}$ and batch size $4$, and train each model for
$50$ epochs. The learning rate is divided by $2$ every $10$ epochs, and the
last checkpoint after the 50th epoch is used for testing. The network width is
set to $32$ channels. For CDD-11 testing, outputs are denormalized to the
$[0,1]$ range and compared with aligned clear targets. For RTTS testing, the
same trained model is directly applied to real hazy images without fine tuning.

\subsection{Main Comparisons}
We compare FreePath with representative dehazing and restoration baselines,
including GCANet~\cite{chen2019gated}, ROP~\cite{liu2022rank},
AirNet~\cite{li2022all}, PDDA~\cite{zhang2026pdda}, and
IHDCP~\cite{liu2026ihdcp}. Table~\ref{tab:main} aligns the quantitative
evidence with the two evaluation regimes: the left group reports full reference
metrics on paired CDD-11 images, and the last column reports no reference
quality on RTTS. FreePath achieves the best CDD-11 scores across PSNR, SSIM,
FSIM, VSI, LPIPS, and GMSD. It also obtains the highest RTTS CLIPIQA score.
These results show consistent gains in pixel fidelity, perceptual structure,
and real scene quality within the same architecture.
The metric pattern is consistent with the intended role of FreePath. PSNR and
SSIM reflect improved paired fidelity, FSIM and VSI indicate stronger
structural preservation, and LPIPS and GMSD show reduced perceptual and
gradient distortion. On RTTS, CLIPIQA further suggests that the learned path
state transfers to real haze without paired real scene supervision.

Figure~\ref{fig:visual} gives the corresponding visual evidence. On CDD-11,
FreePath removes the dense veil while preserving natural colors and scene
structures, producing results closest to the aligned clear targets. On RTTS,
where ground truth is unavailable, FreePath better suppresses veiling while
preserving scene structures, matching the quantitative conclusion in
Table~\ref{tab:main}.

\subsection{Ablation Study}
To verify the effectiveness of the proposed architecture, we conduct an
ablation study under the same data, training setting, and testing protocol as
the main experiment. \emph{FreePath w/o Mean Free Path Residual Block} replaces
the proposed residual blocks with path agnostic residual units, reducing PSNR
from $27.608$ to $27.440$. \emph{FreePath w/o State Conservation Integration}
keeps the Mean Free Path Residual Blocks but removes the conservation principle
at the integration stage, causing a larger drop to $26.644$ PSNR and $0.983$
SSIM. The first variant verifies that the learned path and collision fields
provide useful guidance beyond standard residual transformation, while the
second shows that path states must be preserved during deep aggregation. The
two results jointly support the central claim of neural mean free path
restoration.

\begin{table}[t]
    \centering
    \caption{Ablation studies on CDD-11. Results are mean $\pm$ standard
    deviation under the same training and testing protocol.}
    \label{tab:ablation}
    \vspace{5pt}
    \setlength{\tabcolsep}{1pt}
    \begin{tabular}{@{}lcc@{}}
        \toprule
        Method & PSNR $\uparrow$ & SSIM $\uparrow$ \\
        \midrule
        FreePath w/o MFP Block & 27.440$\pm$4.600 & 0.985$\pm$0.012 \\
        FreePath w/o State Conservation & 26.644$\pm$4.822 & 0.983$\pm$0.013 \\
        FreePath & \textbf{27.608$\pm$4.549} & \textbf{0.986$\pm$0.012} \\
        \bottomrule
    \end{tabular}
    \vspace{-5pt}
\end{table}

\section{Conclusion}
\label{sec:conclusion}

We presented FreePath, an image dehazing framework built around neural mean
free path restoration. Instead of explicitly estimating atmospheric light or
transmission, FreePath learns feature path states inside a residual encoder and
decoder. The Mean Free Path Residual Block performs path sensitive correction,
while FreePath State Conservation Integration preserves restored path states
during deep aggregation. Experiments on paired CDD-11 haze images and real RTTS
scenes show strong performance under full reference and no reference
evaluations, together with consistent visual improvements over representative
baselines.
Ablation results further confirm that path guided correction and state
preserving aggregation are both necessary for effective neural mean free path
restoration.
These results indicate that mean free path is not merely a naming device, but
the organizing variable that connects residual correction and state
conservation into one restoration process.

\clearpage
{\small
\bibliographystyle{IEEEbib}
\bibliography{strings,refs}
}
\end{document}